\documentclass[9pt]{article}
\usepackage{spconf,amsmath,graphicx}
\usepackage{breqn}
\usepackage{tabularx}

\title{Crime Incidents Embedding Using Restricted Boltzmann Machines}
\name{Shixiang Zhu, Yao Xie\thanks{Thanks to the Atlanta Police Foundation for funding. This project is a collaboration with the Atlanta Police Department. The authors would like to thank Mr. Frank Ruben, Lieutenant David Wilson, Major John Quigley at the Atlanta Police Department for technical support for obtaining data and helpful discussions for problem formulation. Also thank Ms. Debra Lam at the Georgia Tech Institute of People and Technology for support.}}
\address{ H. Milton Stewart School of Industrial and Systems Engineering \\
Georgia Institute of Technology \\
Atlanta, GA, USA}

\begin{document}
%\ninept
%
\maketitle
\begin{abstract}
We present a new approach for detecting related crime series, by unsupervised learning of the latent feature embeddings from narratives of crime record via the Gaussian-Bernoulli Restricted Boltzmann Machine (GBRBM). 
This is a drastically different approach from prior work on crime analysis, which typically considers only time and location and at most category information. After the embedding, related cases are closer to each other in the Euclidean feature space, and the unrelated cases are far apart, which is a good property can enable subsequent analysis such as detection and clustering of related cases. 
Experiments over several series of related crime incidents hand labeled by the Atlanta Police Department reveal the promise of our embedding methods.

\end{abstract}
\begin{keywords}
Unsupervised learning, crime data analysis, feature embeddings, neural networks
\end{keywords}
\section{Introduction}
\label{sec:intro}

A fundamental and one of the most challenging tasks in crime analysis is to find related {\it crime series} \cite{3}, which are committed by the same individual or group. Such series of crimes follow a so-called modus operandi (M.O.), for instance, some criminals always break into houses in the late afternoon from backdoor to steal jewels. Finding crime series based on M.O. critically depend on extracting informative features for crime incidents \cite{3}, which is usually done by the human, however, this is not scalable to larger and ever-growing crime data set. For instance, in the City of Atlanta, from the year 2013 to 2017, there are a total of 1,096,961 cases with over 800 categories. 
%finding good representations for each of crime series as their features, thereby predicting or detecting crime series can be easily done. 

Crime incident reports (a.k.a. police reports) are a large source of data that contains rich information for detecting related crime series, which somehow has not been tapped. Each incident has a unique police report, which contains the time, location (latitude and longitude), and last but the not the least, the free-text narratives entered by police officers. According to the crime analysts, the free-text narrative contains the most useful information form their investigation, but there has yet been a tool to automatically extract useful features and information from the free-text narratives, since the narratives are very noisy and unstructured, written by different police officers, and are sometimes incomplete English sentences since they are written in a haste. Currently, crime analysts identify crime series by hand.

%Traditionally, a lot of studies are using spatial-temporal information to model crime series (hotspot prediction \cite{1}). To locate crime patterns, not only do the analysts rely strongly on large amounts of indirectly related information \cite{2}, but also these kinds of methods usually work not very well in the real cases since the crimes are predicted to occur completely based on their occurrence statistics in the past. A better approach like defining modus operandi (M.O.) for characterizing the crime series \cite{3} takes the concrete methods how each crime was committed into account. However, to determine the M.O., it is required to collect very fine-grained detailed information about past crimes, highlights an emphasis on human collaboration within the analysis pipeline, which is costly when it deals with big crime data and heavily depended on the experience of the police detectives.

% The major contributions of this paper includes: 

% \textbf{1}. try to utilize the narratives of crime records to model the crime pattern. 

% \textbf{2}. a novel design of raw feature extraction of narratives of crimes. 

% \textbf{3}. propose a novel idea of learning embeddings of crime records for crime analysis. 

% \textbf{4}. using Gaussian-Bernoulli Restricted Boltzmann Machines (GBRBMs) to learn high-quality feature embeddings from huge amount of criminal narratives without supervision. 

In this paper, we propose a new approach for detecting related crime series that are usually committed by a same group of suspects. This method can directly process the free-text narratives of the police reports, and map them into a feature vector space that automatically captures the similarity of incidents. The main idea is to map the raw feature extracted from the narratives using standard NLP models (such as the bag-of-words model), into latent feature vector space, using Gaussian-Bernoulli Restricted Boltzmann Machines (GBRBMs). The GBRBM is trained from a large number of data without supervision. After training, GBRBM embeds the crime incidents to capture their similarity by vicinity in the Euclidean space. Our work is inspired by the idea of word embeddings \cite{5}, we similarly assume We validate the effectiveness of our method over several series of related crime incidents hand labeled by the Atlanta Police Department.

%for crime analysis that brings together the essential NLP approach and can be used for learning high-quality feature embeddings from huge amount of criminal narratives by Gaussian-Bernoulli Restricted Boltzmann Machines (GBRBMs) without supervision. 

\vspace{0.1in}
\noindent{\bf Relation to prior work}.
A seminal work \cite{3} uses subspace clustering to find crime series and has achieved good performance. However, \cite{3} requires clean features that are entered by the human for each incident. For the larger scale of police report data, there are not many clean hand-entered features. For such dataset, it is highly desirable to be able to directly work with the free-text narratives and find hidden correlations of the crime series.  

A recent work \cite{4} explores the possibilities of using natural language processing tools for crime pattern detection, including Latent Dirichlet Allocation (LDA) and Latent Semantic Analysis (LSA). In certain cases (as we observed in our experiment in Section 3), embedding via GBRBM can be a better approach than LDA, because GBRBM can capture more subtle distinctions between different narratives. This is possibly due to that the structure of RBM directly captures the hidden correlation between incidents via nonlinear structures. We also observe that the narratives usually contain professional but limited "police vocabulary" and has distinct but similar writing styles. This may explain why learning good vector representations is possible, as we have observed from experimenting with millions of crime records in the Atlanta Police Department database.

Another line of research from crime analysis and predictive policing focus on the so-called {\it hotspot prediction} (see, e.g., \cite{1,2}), which has achieved a lot of success in modeling the dynamics of crimes over space and time. The hotspot prediction aims to model the excitation relationships between crime incidents occurred at different space and time. The idea is that certain types of crimes (such as gang crimes) have a triggering effect: an event happens at certain location and time may trigger future incidents at similar locations in the futures. The hotspot prediction typically only use time, location (and sometimes category) information of the incidents.

%The major contribution of this work are the ideas of leveraging narratives in crime records and finding their latent embeddings, there formed drastic contrast from the earlier work. It is motivated by the word embeddings \cite{5}, but takes the whole concept of learning embedded representations from the words to the level of documents. Our work is a drastic contrast from previous research, the narratives of the crime reports, which usually contains the most critical information of the crimes, have been somehow overlooked since the narratives are too noisy and unstructured to engineer their features effectively. 

\section{Embedding using GBRBM}
\label{sec:methodology}

%\subsection{Overview}
%\label{ssec:overview}

Our dataset is provided by the Atlanta Police Department (APD), which consist of all crime incidents from 2013 to 2017, with 1,096,961 cases in over 800 different categories. The records are unlabeled and naive clustering will not resolve them into related crime series. The latent feature embedding algorithm that we describe below capture the critical information of the crime and criminal correlations would be very helpful no matter for further classification or clustering of the crime cases in the absence of label information. 

%The feature embeddings of the crime cases computed using GBRBMs are very interesting since the learned representations explicitly encode many criminal regularities and patterns. Using the proposed method where the encoding process is performed on the raw features of the narratives, it shows the feature embeddings are the high-quality representations for the each of the narratives of the crimes and are also more compact than the raw features which would be very memory-friendly to the follow up classification or clustering.

\begin{figure}
\centering
\centerline{\includegraphics[width=1\linewidth]{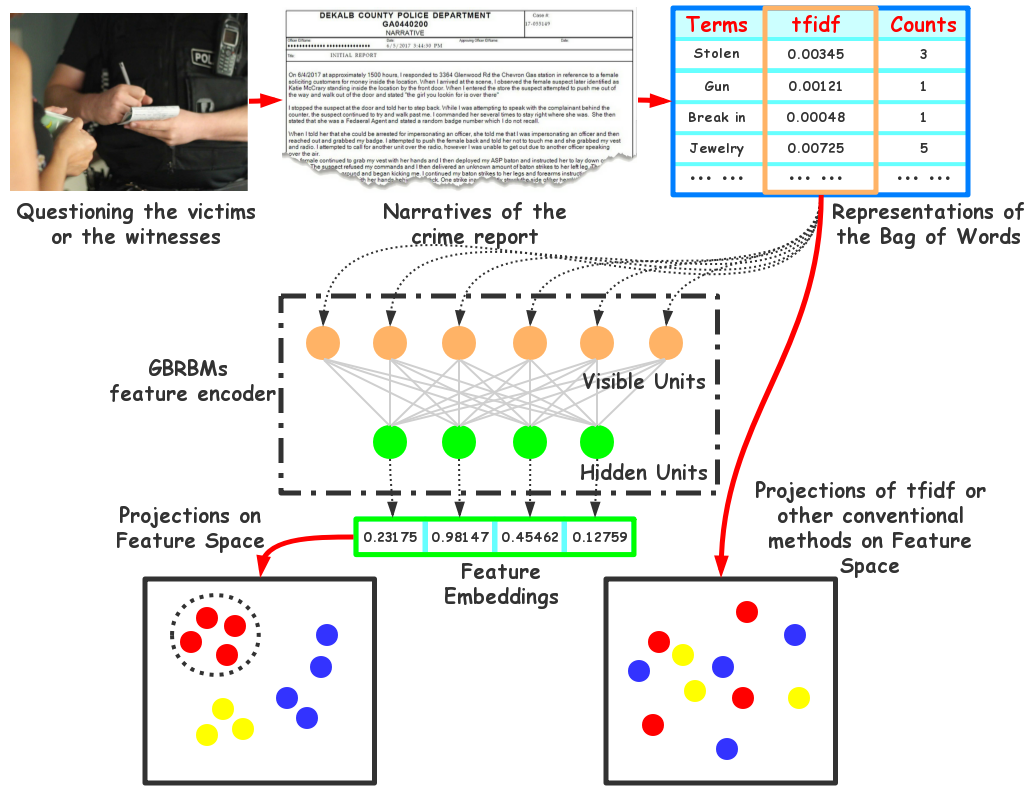}}
\caption{An overview of the essential idea of our method.}
\label{fig:illustration}
\end{figure}

The flowchart of the embedding algorithm is shown in Fig. \ref{fig:illustration}. On a high level, we first generate raw features using standard natural language processing tools, such as term-frequency inverse-document-frequency (TF-IDF) for each incident. %When defining terms, we combine a list of 168 key-words provided by the police officers, which are words deemed important according to their experiences, with augmented key words that are extracted from the police report corpus. 
The core of the algorithm is the GBRBM with the input being the TF-IDF of reports. We train the GBRBM by maximizing the likelihood function (defined in terms of the energy function) for the occurrence of word terms in a police report, and in the end, use the output latent variables as the embedded features for each corresponding crime incident. The embedded features have a nice property that related cases have features in the vicinity in the Euclidean space. 

%All the narratives that we have will be used for the unsupervised learning of the GBRBMs before computing the feature embeddings. Eventually, the model shows a surprisingly good performance on learning the key structure of the narratives compared with other traditional NLP methods. We will give some detailed examples in section 3.

\begin{figure}[h!]
\centering\vspace{-0.1in}
\centerline{\includegraphics[width=1\linewidth]{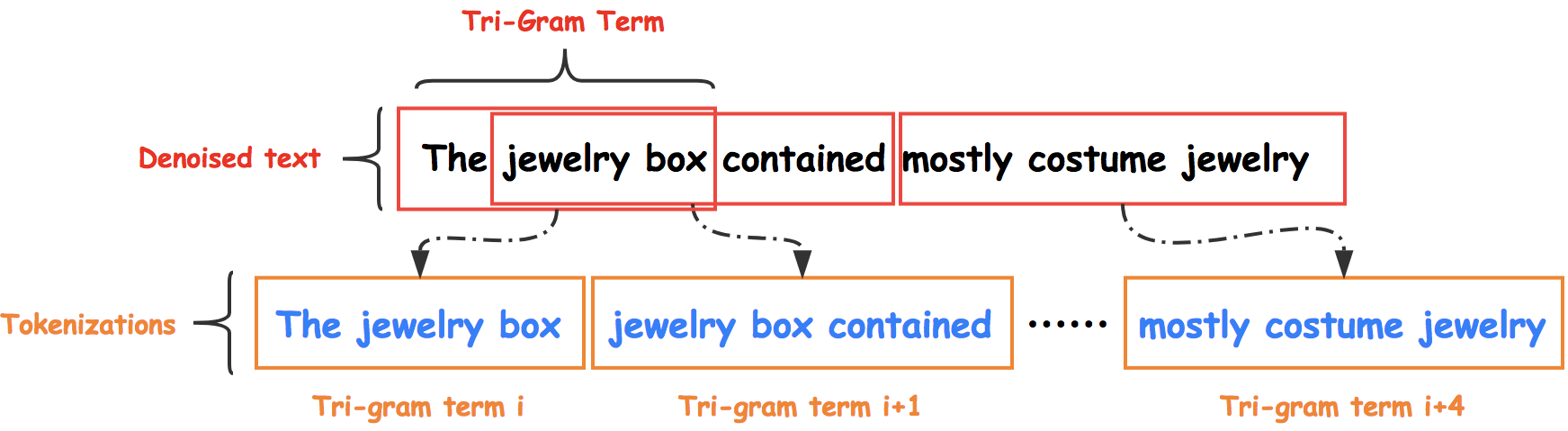}}
\caption{An example of Tri-Gram.}
\label{fig:trigram}
\vspace{-0.2in}
\end{figure}

\subsection{Raw Feature Extraction}
\label{ssec:feature_design}

The free-text narratives are highly unstructured data, meanwhile they also consist typos, irrelevant words or phrases. Our raw feature extraction is designed to be robust to these issues. Several key steps are as follows:
%
%we design several key preprocessing procedures to reorganize the text into a more friendly form for feature modeling before the modeling:
%
\textbf{Data cleaning}: we normalize the text to lower-cases so that the distinction between "The" and "the" are ignored; also remove stop-words, independent punctuation, low-frequency terms (low TF) and the terms that appeared in most of the documents (high IDF).
\textbf{Tokenization}: for each narrative of the crime, the text needs to be tokenized into a list of word-level tri-gram terms, as shown in Fig.\ref{fig:trigram}. As a matter of experience, unigram and bigram loss too much context information, while the four-gram or higher gram secures only tiny gains while making the feature vector become too long and increasing the model complexity and training is more time-consuming.
\textbf{Bag of Words (BoW)}: BoW is a simplifying, orderless document representation commonly used in NLP. In this representation, each document is represented by one vector where each element means the occurrence in association with a specific term. As a result, the entire corpus can be converted to a term-document matrix and a dictionary that keeps the mapping between the terms and their ids.
\textbf{Term Frequency-Inverse Document Frequency (TF-IDF)} is a conventional method for extracting feature vectors from the term-document matrix to de-emphasize frequent words. In \cite{10}, TF-IDF weighting scheme has been used to reduce the impact of the terms that appeared in most of the documents, which means that they have weak discrimination capability across documents.

\subsection{Model Architectures}
\label{ssec:model_architectures}

GBRBM is a type of neural networks and it can be viewed as a probabilistic graphical model. GBRBM is a powerful model for the complex joint distribution of real-valued visible variables and binary valued hidden variables \cite{6}. Here, we consider a standard GBRBM, whose architecture is shown in Fig.\ref{fig:illustration}, 
where there are $n$ hidden variables as embeddings and $m$ visible variables that will be input by tokenized word terms. The weights, represented as a matrix $W=(w_{ij})$, associate the hidden variable $h_j$ and visible variable $v_i$. There are also bias weights $c_j$ for the hidden variable and $b_i$ for the visible variable. The learned weights and biases define a Gibbs probability distribution over all possible input data via the energy function, denoted as $E(\mathbf v, \mathbf h)$. The energy function for the joint configuration $(\mathbf v, \mathbf h)$ of the visible and hidden units is defined \cite{7}
\[
E(v, h) = - \sum_{i,j} w_{ij} h_j \frac{v_i}{\sigma_j} - \sum_i \frac{(v_i - b_i)^2}{2 \sigma^2} - \sum_{j} h_j c_j.
\]
Note that the first term of the energy function captures the joint pattern of the hidden and visible variables, and the second and the third terms capture the linear effect of both the hidden and the visible variables. 
A nice structure of the RBM is that the joint distributions of the visible and hidden variables, conditioned on each other, are mutually independent
\[
p(v|h)= \prod_{i=1}^{m} p(v_i |h), \quad
 p(h|v) = \prod_{j=1}^{n} p(h_j |v).
\]
Moreover, the conditional distribution of $v_i |h$ is a normal random variable $\mathcal{N}( b_i + \sigma_i \sum_{j=1}^{n} {w_{ij}h_j}, \sigma_i^2)$  \[p(v_i=v|h)=\frac{1}{\sigma_i \sqrt{2\pi}} \cdot e^{-\frac{1}{2\sigma_i^2} (v_i - b_i - \sigma_i \sum_{j=1}^n w_{ij} h_j)^2},\] and the conditional distribution of $h_i |v$ is Bernoulli variable  \[p(h_j=1|v)=\sigma(\sum_{i=1}^{m} {w_{ij}\frac{v_i}{\sigma_i} + c_j}).\] where $\sigma$ is a sigmoid function. Due to this property, it is easily to sample visible or hidden variables via Gibbs sampling \cite{11} in just two steps: sampling a new state $h$ for the hidden units based on $p(h|v)$ and sampling a state $v$ for the visible layer based on $p(v|h)$. %Due to this special structure, the computation for GBRBM can be done efficiently: one can sample the visible variables or the hidden variables via Gibbs sampling \cite{11}. 
The sampling procedure is essential to perform model estimation. %, as we explain below. 

%\begin{figure}[h]
%\centering
%\centerline{\includegraphics[width=0.6\linewidth]{img/gbrbm}}
%\caption{The architecture of the GBRBMs.}
%\label{fig:rbm}
%\vspace{-0.1in}
%\end{figure}

\subsection{Model Estimation}
\label{ssec:model_estimation}

We follow the standard training approach for GBRBM. The training objective of the GBRBM is to maximize a likelihood function, which is defined via the energy function. The training result, somehow, in the end, converges to representations such that related cases tend to be close to each other in Euclidean space. More formally, given a set of training narratives $\mathbf V = \{v^{(1)}, v^{(2)}, v^{(3)}, \ldots, v^{(N)} \}$, the objective of the model is to maximize the average log likelihood given by 
$\log\mathcal L (\theta | V) = \sum_{i=1}^{N} \log p (v^{(i)}|\theta), $
where the marginal distribution is given by
\[p(v) = \sum_{h} {p(v, h)} = \frac {\sum_{h} {e^{-E(v, h)}}}{\sum_{v,h}{e^{-E(v, h)}}}.\]
%where $h$ is the latent representation of $v$ that we want to estimate, $p(v,h)$ is the joint probability distribution of every possible pair of $(v,h)$, and $p(v)$ is the marginal distribution of $v$. 
Note that the number of possible values of $h$ vectors is exponential in the number of hidden variables, so in practice, one usually performs sampling approach to calculate the sum approximately. 

Directly obtaining unbiased estimates of the log-likelihood gradient using MCMC methods typically requires many sampling steps. In training, we adopt the $k$-step contrastive divergence (CD-$k$) approach, which is an approach to approximate the gradient in training GRB via gradient descent \cite{8}. The main idea is to approximate the gradient of the log-likelihood with respect to $\theta$ for one training pattern $v^{(0)}$ as
\vspace{-0.1in}
\begin{equation*}
\begin{split}
{\rm CD}_k (\theta, v^{(0)}) 
&= -\sum_{h} {p(h|v^{(0)}) \frac {\partial E(v^{(0)}, h)}{\partial \theta}}\\
& + \sum_h {p(h|v^{(k)}) \frac {\partial E(v^{(k)}, h)}{\partial \theta} }.
\end{split}
\end{equation*} The Gibbs chain is initialized with a training example $v^{(0)}$ of the training set and yields the sample $v^{(k)}$ after $k$ steps. Each step $t$ consists of sampling $h^{(t)}$ from $p(h|v^{(t)})$ and sampling $v^{(t+1)}$ from $p(v|h^{(t)})$ subsequently. The iterations are repeated until certain empirical convergence has achieved.

%Typically how many samples are needed for MCMC to converge?

\section{Results}
\label{sec:results}

To test the performance of our embedding method, we devise a comprehensive test dataset. The dataset contains five hand-labeled crime series that were identified as committed by five individual arrestees, and 441 randomly selected irrelevant crime cases. Details of the test data are given in Table~\ref{tab:test_data}. 

Ideally, we hope that the crime records which were committed by the same arrestee tend to be closed to each other in the embedded feature space. However, it is not so easy to show the distance between two crime cases directly without dimensionality reduction on their feature embeddings. For visualization purposes, we apply two-dimensional t-distributed stochastic neighbor embeddings (t-SNE) \cite{9} to the feature embeddings, to convert the high-dimensional feature vectors into a matrix of pairwise similarities. t-SNE is capable of capturing local structure of the high-dimensional data, while also revealing global structures such as the presence of clusters at several scales \cite{9}.

\begin{table}[h!]
\centering
\caption{Details of the test data} \label{tab:test_data}
\vspace{.1in}
\begin{tabular}{|l|l|l|}
\hline
\textbf{Id} & \textbf{Number} & \textbf{Category} \\ \hline
Crime Series 1 & 8 & \textit{Robbery at Residence} \\
Crime Series 2 & 7 & \textit{Robbery at Gas Station} \\
Crime Series 3 & 4 & \textit{Pedestrian Robbery} \\
Crime Series 4 & 15 & \textit{Attempt Auto Theft} \\
Crime Series 5 & 22 & \textit{Burglary} \\
Random Cases & 441 & \textit{Over 89 Categories}\\\hline
Total & 497 & 
 \\\hline
\end{tabular}
\end{table}

\begin{table}[h!]
\centering
\caption{The comparison of the training time.} \label{tab:training_time}
\vspace{.1in}
\begin{tabular}{|l|l|}
\hline
\textbf{Methods} & \textbf{Training Time} \\ \hline
GBRBMs with 1000 units & \textit{$\sim$ 2 mins} \\
GBRBMs with 2000 units & \textit{$\sim$ 3 mins} \\
GBRBMs with 5000 units & \textit{$\sim$ 7 mins} \\
LDA with 1000 topics & \textit{$\sim$ 5 mins}\\\hline
\end{tabular}
\vspace{-0.1in}
\end{table}

\begin{figure}[h!]
\vspace{-0.1in}
\begin{center}
\begin{tabular}{c}
\includegraphics[width=0.67\linewidth]{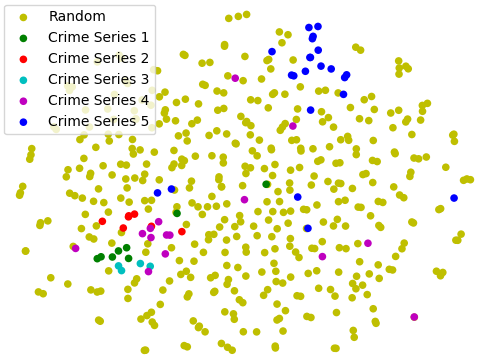}\\
(a) GBRBMs with 1000 hidden nodes\\
\includegraphics[width=0.67\linewidth]{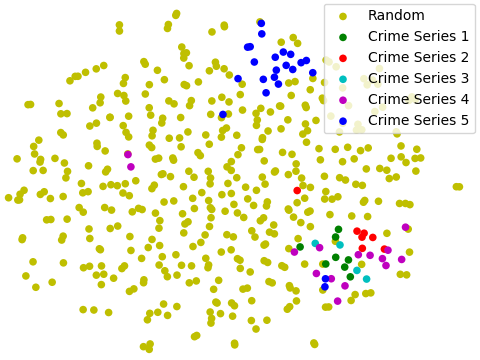}\\
(b) GBRBMs with 2000 hidden nodes\\
\includegraphics[width=0.67\linewidth]{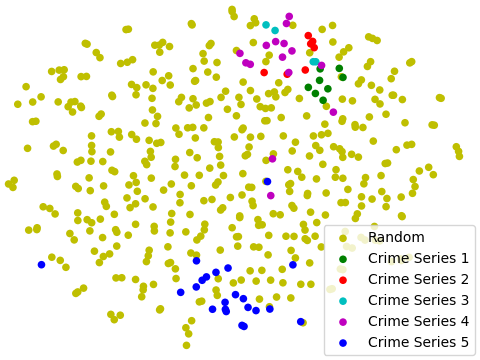}\\
(c) GBRBMs with 5000 hidden nodes\\
\includegraphics[width=0.67\linewidth]{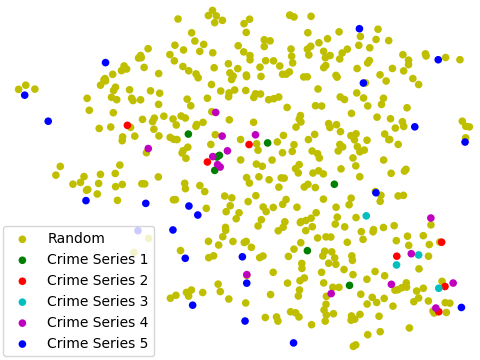}\\
(d) LDA with 1000 topics\\
\end{tabular}
\end{center}
\caption{Visualization of the projections of different embeddings on the 2D t-SNE space.}
\label{fig:visualizaton}
\vspace{-0.2in}
\end{figure}

%
%\subsection{Examples}
%\label{ssec:examples}

In our experiments, the basic parameters for the GBRBM are as follows: the size of the visible layer is fixed to 9863, which is determined by the size of the dictionary. We tried 3 different sizes of the hidden layer for testing the model   performance, with 1000, 2000 and 5000 hidden units respectively. In the training stage, the learning rate is 0.05, the batch size is 20, and the number of epochs at fine tune periods is 30. We adopt the Stochastic Gradient Descent (SGD) optimizer \cite{Duchi2011} to optimize the loss function.

First, we study the effect when increasing the number of hidden variables in GBRBM. This will lead to different dimensions of the feature embeddings. Fig.\ref{fig:visualizaton} (a) shows that the embeddings with 1000 units can successfully map crime series 1, 2, 3, 5 to clusters, which separate them out from random cases. In particular, for crime series 1, 2, 3, the embeddings of same crime series gather closely at some local regions. The clustering does not work quite well for crime series 4. When we increase the number of the hidden units to 2000, the performance for crime series 4 become much better as shown in Fig.\ref{fig:visualizaton} (b) and (c). The performance of the GBRBM does not seem to have significant further  improvement when we further increase the number of the hidden variables (Fig.\ref{fig:visualizaton} (c)).

Second, we compare the performance of GBRBM with Latent Dirichlet Allocation (LDA) \cite{Blei2003} on the same test data set. Fig.\ref{fig:visualizaton} shows four instances, which are the projection of the embeddings via GBRBM and LDA topic modeling on a 2D t-SNE space. We implement a LDA with 1000 latent topics. It turns out that the LDA does not map crime series into clusters: they are scattered randomly in the feature space without any obvious patterns. 

The embedding can be computed efficiently. As reported in Table~\ref{tab:training_time}, the training time of reaching the convergence precision for 497 cases are around minutes. This also shows the GBRBM with less than 2000 hidden units have an advantage over the LDA in terms of the training time.

\vspace{-0.1in}
\section{Conclusion}
\label{sec:conclusion}

We have presented a novel approach for detecting crime series that are related, using embedding found by the Gaussian-Bernoulli Restricted Boltzmann Machine (GBRBM). The GBRBM tends to map related cases (that share certain correlation in the raw feature) into features that are in the vicinity in the Euclidean space. Our methods demonstrate very promising results on real police data and demonstrated that the feature embeddings can have advantages over the conventional text processing methods on detecting crime series in certain cases. Ongoing work is to develop an online crime series detection algorithm based on the embedded features. 

% embeddings from the narratives of the crime reports. By taking input from the preprocessed narratives, a well-trained GBRBMs as an encoder can produce compact and high-quality representations that could be used in further analysis. Concrete examples have been demonstrated that the feature embeddings have obviously advantages over the conventional text processing methods on detecting the pattern of the crimes. 

%\section*{Acknowledgement}
%\label{sec:acknowledgement}
%
%The authors would like to thank Mr. Frank Ruben, Lieutenant David Wilson, Major John Quigley at the Atlanta Police Department for technical support for obtaining data and helpful discussions for problem formulation. Also thank Ms. Debra Lam at the Georgia Tech Institute of People and Technology for support.

\clearpage

\bibliographystyle{ieeetr}
\bibliography{refs}

\end{document}